\documentclass{article}

\usepackage{arxiv}

\usepackage[T1]{fontenc}
\usepackage[utf8x]{inputenc}
\usepackage{hyperref}       
\usepackage{url}            
\usepackage{booktabs}       
\usepackage{amsfonts}       
\usepackage{nicefrac}       
\usepackage{microtype}      
\usepackage{lipsum}		
\usepackage{graphicx}
\usepackage{natbib}
\usepackage{doi}

\title{Identifying the style by a qualified reader on a short fragment of generated poetry}

\author{\href{https://orcid.org/0000-0002-9099-0436}{\includegraphics[scale=0.06]{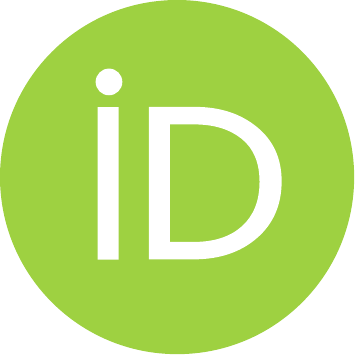}\hspace{1mm}Boris Orekhov}\footnote{webpage: \href{http://nevmenandr.net/bo.php}{http://nevmenandr.net/bo.php}, 	alternative email: nevmenandr@gmail.com} \\
	School of Linguistics\\
	HSE University,\\
	Institute of Russian Literature (Pushkin House)\\
	Russian Academy of Sciences \\
	\texttt{borekhov@hse.ru} 
}

\renewcommand{\shorttitle}{Identifying the style by a qualified reader}

\hypersetup{
pdftitle={Identifying the style by a qualified reader on a short fragment of generated poetry},
pdfsubject={cs.LG, cs.CL},
pdfauthor={Boris Orekhov},
pdfkeywords={computer generated poetry, style, evaluation},
}

\begin{document}
\maketitle

\begin{abstract}
	Style is an important concept in today's challenges in natural language generating. After the success in the field of image style transfer, the task of text style transfer became actual and attractive. Researchers are also interested in the tasks of style reproducing in generation of the poetic text. Evaluation of style reproducing in natural poetry generation remains a problem. I used 3 character-based LSTM-models to work with style reproducing assessment. All three models were trained on the corpus of texts by famous Russian-speaking poets. Samples were shown to the assessors and 4 answer options were offered, the style of which poet this sample reproduces. In addition, the assessors were asked how well they were familiar with the work of the poet they had named. Students studying history of literature were the assessors, 94 answers were received. It has appeared that accuracy of definition of style increases if the assessor can quote the poet by heart. Each model showed at least 0.7 macro-average accuracy. The experiment showed that it is better to involve a professional rather than a naive reader in the evaluation of style in the tasks of poetry generation, while lstm models are good at reproducing the style of Russian poets even on a limited training corpus.
\end{abstract}


\section{Introduction}

Style is an important concept in today’s challenges in natural language generation. When and if we want to see how style is revealed through text generation, we should approach style as a combination of formal features (ngrams, words, character distribution). There is an extensive literature that covers a range of approaches to studying formal features in fiction and poetry. A number of papers look at the distribution of most frequent words \citep{burrows2002delta, rybicki2011deeper}. Another group of works studies word sequences \citep{pawlowski1998series}. 

After success in the field of image style transfer \citep{gatys2016image}, recent literature considers text style transfer as changes in textual grammar in an attempt to see significant features that can distinguish different styles \citep{kabbara2016stylistic, li2018delete, yang2018unsupervised}. However, the idea of style as a combination of grammatical categories \citep{fu2018style} seems disputable as there is no agreement in literature what features or their combinations make style of individual authors in fiction or poetry. The definitions of style emphasize the vagueness of the concept of style as “refers to the way in which language is used in a given context, by a given person, for a given purpose, and so on” \citep{short2013style}. We focus “not just on what is said, but on how it is said, and why a text may be shaped in a particular way” \citep{youdale2019using}. Moreover, style is “what is unique to a text” \citep{boase2017stylistics}. However, \citep{orekhov2020neural} propose methods that identify similar formal features in texts generated by neural networks and texts used as a training dataset.

Developers are also interested in the tasks of style reproducing in generation the poetic text. As it said by the researchers: “Poetry generation is becoming a mature research field, which is confirmed by several works that go beyond the production and exhibition of a few interesting poems that, to some extent, match the target goals.” \cite{oliveira2017survey} 

Competitions in poetry generation targeted at NLP researchers contribute to solving poetry generation tasks. This is important for improving personal assistants’ linguistic abilities and enlarging their capacities to understand and produce a variety of linguistic styles and formats. 

One of the theoretical problems that concern us in this area is the problem of enough text to manifest style. Intuitively, it seems that style will not be able to manifest itself in a short text. We will test this with an example of a four-line text.

Evaluation of style reproduction remains a problem. In a 2018 competition, \href{http://web.archive.org/web/20180907001018/https://classic.sberbank.ai/description}{Sberbank}  relied on the subjective opinion of assessors, who had to say whether the presented text corresponded to the style of a presented poet. There were also other works  in which the task was, to evaluate how well the style of the original author was reproduced \citep{potash2015ghostwriter}. In my opinion, such assessment methods overlook important features that I want to pay attention to.

\section{Experiment design}
\label{sec:design}

How do you check that neural networks can really reproduce a specific style? The simplest option is to show a passage of text to an expert and ask him to answer the question as to whether this passage is similar to the work of some specific author. But, as is well known, any complicated problem has a simple, obvious and completely wrong solution.

Where can we find  such an answer? For example, it may reflect the respondent’s sympathy to the experimenter. If the respondent is favorable, he or she will answer “yes” to please the experimenter, and “no” if he or she wants to be hard on the experimenter.

This answer may also reflect the degree of familiarity of the respondent with the poet he or she is asked about. If the respondent is not familiar with the poet being asked about, his or her grade will be poor.

In the Sberbank competition mentioned above, the matching of style was assessed by specially hired people: they were shown the texts created by the generator, and people rated those texts. I can’t officially confirm it, but according to rumors, at some point there was an unexpected failure: the generator began to produce not its  own texts, but the original texts of Russian poets. Assessors said that these texts do not match the style of referenced poets. But even if it’s just an anecdote created by someone, how are we protected from it really taking place? With this  type  of  experiment design, there is no way.

First of all, it is necessary for the evaluator not to know what style the machine is trying to reproduce. Ideally, he should give the name of the author himself. But since Russian poetry is rich in notable names, the respondent may simply not have time to think about the right person during the experiment, especially if the text presented to him is short. In a small passage, the author’s signal may be muted and it is inconvenient to show a long text during the experiment. The compromise will be a closed list of names, one of which belongs to the very poet, on whose texts the model was trained.

The second is to find out how familiar the evaluator is to the author about whom he is asked. For this purpose, you can, for example, ask him an additional question. Obviously, the respondent, who knows the poet well, would not be the hero of the anecdote about the Sberbank competition: he would simply recognize in the lines shown to him, verses of a famous poet.

Thirdly, is that not everyone would be able to take part in the experiment., It is necessary to involve those who are attentive to the text. Style is not what the text says, but how it is said. We have to admit that this is how only professional readers look at the word. That is, literary scholars and philologists, whose professional competence is to pay attention to how things are  said.

\section{Data and questions}
\label{sec:data}

\subsection{Data overview}

I’ve trained three character-based LSTM neural network models to work with a style reproducing assessment. All three models were trained on the corpus of texts by famous Russian-speaking poets; Nikolay Nekrasov (1,200,000 characters, 191,000 words, 54,000 lines in input, 5 layers, network size 512, loss value = 0.9298), Osip Mandelshtam (265,000 characters, 39,000 words, 14,000 lines in input, 7 layers, network size 512, loss value = 1.0705), and the early works of  Boris Pasternak (316,000 characters, 50,000 words, 7,000 lines in input, 7 layers, network size 512, loss value = 1.1266).

Respondents were students of Bachelor’s and Master’s degree programs at the Higher School of Economics, specializing in the history of literature. The questions were answered by 94 people. It seemed to me to be important to involve only qualified people in this experiment, who really know what a literary poetic style is. It would be even better to involve teachers, professors specializing in the history of literature in the experiment. But the time of such specialists is very expensive, and with them I would never get such a mass experiment.

From the samples, I randomly selected four lines of text for each model.

\subsection{Nekrasov model}

Lines from Nekrasov model were:

\begin{verse}
I kartochki ne slyshal. \\
On byl uzh dobryj svet, \\
No kak by mog pribavil \\
Kakoy-to bednogo pokoy.
\end{verse}

See translation:

\begin{verse}
And didn’t hear the card.\\
He was a good light,\\
But how could I have added\\
Some kind of poor peace.
\end{verse}

The text does not rhyme (like other samples of all models). The only word that draws attention to itself is \textit{kartochka} ‘card’. This word is associated with ration cards, which only came into life in the 20th century (Nekrasov is a poet of the 19th century). But in Nekrasov’s train texts there is a word \textit{kartochka}: \textit{Igrala v kartochki do petukhov}, ‘played cards till morning’.

In the questionnaire, 4 possible candidates for authorship of the source text were proposed for this fragment: E. Belov (fictional  poet), N. Nekrasov, M. Kuzmin, P. Vyazemsky. Nekrasov and Vyazemsky are 19th century poets. The respondent will probably choose between these authors if he understands that it is the 19th century that stands behind the stylistics of text. M. Kuzmin is a Russian poet of the early 20th century. Option E. Belov, is needed in order to test the integrity of respondents and their ability to answer questions responsibly, rather than giving random noisy answers. If many respondents choose the option “E. Belov”, it means that they are not serious about the experiment and are not ready to evaluate the text style. At school, the Nekrasov style is usually said to be inconsistent with the early 19th century poetic patterns and seeks to be similar to prose.

\subsection{Mandelshtam model}

Lines from the Mandelshtam model were:

\begin{verse}
Pod derev'ya polnochnogo vozdukha.\\
Na vechnosti v otkaze vernetsya,\\
I nashim novym pustotelym plat'em \\
Na prozrachnoy podkove prosili leta.
\end{verse}

See translation:

\begin{verse}
Under the trees of midnight air.\\
For eternity in rejection will return,\\
And our new hollow dress.\\
They asked for summer on a transparent horseshoe.
\end{verse}

4 candidates for the input corpus authorship were: O. Mandelshtam, A. Akhmatova, M. Tsvetaeva, V. Mayakovsky. O. Mandelshtam and A. Akhmatova were in the same literature group ‘akmeists’, and that forces us to think that their poetic styles are similar. The choice between these two options should be hard. The other poets on this list were creatively active at the same time. Only actual knowledge of their stylistics can help to establish the style of the source text. At the same time, if respondents can understand which author’s style is reproduced here, it will be difficult to consider this result as random.

The word \textit{plat'e} ‘dress’ in accordance with its historical semantics can mean a gender-neutral dress. But for the modern reader it should be associated with a woman’s dress. It is most likely to mislead respondents, as they will think that the author is a woman.

\subsection{Pasternak model}

The lines from early the Pasternak model were:

\begin{verse}
Kak v sumerki mysl',\\
lish' gorod i lyudi byli kak pyl'nik.\\
Mozhet, kak novodorodnyy golos,\\
Odnogo list'ev i podnosit pryada.
\end{verse}

See translation:

\begin{verse}
It’s like a thought at dusk,\\
only the city and people were like a duster.\\
Maybe like a newborn voice,\\
One leaf and a strand.
\end{verse}

The word \textit{novodorodnyj} doesn’t exist. It was created by the neural network because of its character-based nature. Because of this neologism, apart from Pasternak, I put as an option Mayakovsky, a poet who liked to compose new words for his poems. The other two poets were contemporaries of B. Pasternak, S. Esenin and N. Gumilev. Their stylistics were very different from those of Pasternak. But I was not sure that 4 lines would be enough for respondents to identify the poet’s style.

In addition to the text and list of possible authors, this question was included in the questionnaire: “How familiar are you with the work of the poet who was chosen in the previous question?” This question had 5 options: 1) I’ve never heard that name before. 2) I’ve heard of him, but I have a vague idea who he is. 3) I am familiar with this author in general terms, but I can’t quote. 4) I am familiar with this author and can quote a few lines or verses 5) this author is well known to me, I remember many of his poems by heart.

Such a question not only allows one to understand whether the respondent was able to identify the author’s style, but also allows one to understand whether he did it by chance or through a deep acquaintance with the work.

\section{Results}

Respondents were most likely to choose the right option for all three passages (see table \ref{tab:table}). Respondents correctly identified the Nekrasov model in 38 cases (40.4 \%), the Mandelshtam model in 41 cases (43.6 \%), and the Pasternak model in 46 cases (48.9 \%). It is significant that all respondents who named Belov in the first question answered honestly that they had never heard such a name. This means that the quality of answers is high. Respondents did pay attention to the experiment, they did not respond automatically.

\begin{table}[htbp]
	\centering
	\caption{Results}
	\begin{tabular}{@{}ccccc@{}}
		\toprule
		Model       & Correct & Incorrect 1 & Incorrect 2 & Incorrect 3 \\ \midrule
		Nekrasov    & 38      & 8           & 22          & 26          \\
		Mandelshtam & 41      & 33          & 19          & 1           \\
		Pasternak   & 46      & 20          & 18          & 10          \\ \bottomrule
	\end{tabular}
\label{tab:table}
\end{table}

Respondents’ answers can be described as classifier decisions in a multiclass classification. Each right answer is a true positive prediction. Since I’ve been dealing with multiclass classification, just calculating accuracy isn’t enough. There were more than one wrong choice, which means that the value of the correct answer increases. I used the macro-average accuracy \citep{van2013macro}. Macro-average accuracy for Nekrasov case is 0.702, for Mandelshtam case is 0.718, for Pastrnak case is 0.744. As we can see, the value of accuracy is not perfect, but far from being random. Our LSTM network really represents the author’s style, which can be defined even by choosing between the authors who are close by the period and literary position. Now we can check how the degree of acquaintance with the author’s work influences the choice. All levels of acquaintance with the authors presented can be divided into two types: 1) a respondent can quote by heart from the author, and 2) a respondent cannot quote by heart from the author.

If we take only type 2 respondents, the quality of style definition grows significantly. People who can quote the poet’s lines by heart correctly, define Nekrasov’s style in 80.6 \% of cases. That gives us 0.87 macro-average accuracy. Almost everyone who answered "Vyazemsky" cannot quote a single line from this poet. Therefore, the option "Vyazemsky" is almost always a noise. For a qualified reader who really understands the history of literature, the difference between the Nekrasov and Vyazemsky styles is significant. A model trained on Nekrasov’s texts cannot generate texts similar to Vyazemsky.
The quality of the answers in the other two cases does not increase so noticeably, because the answer choices are very close in period and style, but still growing nonetheless (see table \ref{tab:table2}).

\begin{table}[htbp]
	\centering
	\caption{Experts who can quote by heart from the author}
	\begin{tabular}{@{}cccccc@{}}
		\toprule
		Model       & Correct & Incorrect1 & Incorrect2 & Incorrect3 & Macro-average accuracy \\ \midrule
		Nekrasov    & 29      & 0          & 6          & 1          & 0.87                   \\
		Mandelshtam & 35      & 26         & 14         & 1          & 0.73                   \\
		Pasternak   & 35      & 19         & 11         & 9          & 0.74                   \\ \bottomrule
		
	\end{tabular}
\label{tab:table2}
\end{table}

\section{Discussion}

We have once again confirmed that LSTM models can reproduce the style of the train corpus. A closed list of options, of course, limits us in understanding how well the respondent can define the style of the model samples. But the ability of a qualified reader to distinguish the style, even in a situation of choice between similar in epoch and literary position authors, shows that this is not a random property of the model. If this effect is already confirmed, we can assess not the effect itself, but the experts and their influence on the assessment process.

I am aware that for the completeness of the picture, it would be necessary to ask about the degree of acquaintance with the work of not only the author, whom respondents pointed to in the previous question, but with the work of all the poets in the list. But my goal was to make the questionnaire as simple as possible. This is what allowed 94 qualified experts to be involved in the work. At the same time, such a design of this experiment, in which we do not show the respondent a specific text, but refer to his memory and the overall impression of the entire work of the author, seems more consistent with the idea of style, in literary studies.

Of the experts involved 18 \% did not give any correct answer, 41 \% gave only one correct answer, 30 \% gave two correct answers, and 11 \% answered all the questions correctly. This underlines the specialization of experts. If one understands Pasternak’s style well, it is not necessarily that one understands Nekrasov’s style as well.

Obviously, the most frequent correct pair of right answers was Pasternak and Mandelstam (22 times), as they are poets of the same period. The experts who guessed these poets seem to be interested in the poetry of modernism. Nekrasov and Mandelstam were guessed 19 times, Pasternak and Nekrasov were guessed 17 times by the same respondent. Nekrasov and Pasternak are indeed very far apart by their style.

At the same time, it is obvious that there are 10 more  successful experts who have the best performance in style definition. However, these experts do not correlate with knowledge of poetry by heart. Only 4 out of 10 respondents knew poetry by heart in all three cases.

\section{Discussion}

The context of style is all the poet’s work or at least his most significant poems, which remain in the memory of the reader. It is not enough to present one line to the respondent as a golden standard and ask him if the generated text, by style, is similar to what he sees in this line. Usually the whole poet’s work, or his most famous lines, are stored in the expert’s memory.

The research shows that it is not reasonable to involve a naive reader in style assessment tasks. But short texts like 4 lines are enough to evaluate the reproducibility of the style in a computer generated poetry.

How familiar  the expert is with the poet whose style should be evaluated in samples of the model, can be ranked by whether he knows the lines from this poet by heart.

\section*{Acknowledgements}
I am grateful to Inna Kizhner, and Julia Flanders for their comments and suggestions. I am also grateful to Yana Linkova, who made this research possible. %

\bibliographystyle{unsrtnat}
\bibliography{references}

\end{document}